\documentclass[10pt,twocolumn,letterpaper]{article}

\usepackage{cvpr}
\usepackage{times}
\usepackage{epsfig}
\usepackage{graphicx}
\usepackage{amsmath}
\usepackage{amssymb}

\DeclareMathOperator*{\argmin}{argmin}

\usepackage{multirow}
\usepackage{bm}
\usepackage{float}
\usepackage{algorithm}
\usepackage{rotating}
\usepackage{array}
\usepackage{url}

\usepackage[pagebackref=true,breaklinks=true,letterpaper=true,colorlinks,bookmarks=false]{hyperref}

\cvprfinalcopy

\begin{document}

\title{EpicFlow: Edge-Preserving Interpolation of Correspondences for Optical Flow}

\author{Jerome Revaud$^{a}$ \qquad Philippe Weinzaepfel$^{a}$ \qquad Zaid Harchaoui$^{a,b}$ \qquad Cordelia Schmid$^{a}$ \\
$^{a}$ Inria\thanks{LEAR team, Inria Grenoble Rhone-Alpes, Laboratoire Jean Kuntzmann, CNRS, Univ. Grenoble Alpes, France.} \qquad \qquad $^{b}$ NYU \\
{\tt\small firstname.lastname@inria.fr}}

\maketitle
\thispagestyle{empty}

\begin{abstract}
We propose a novel approach for optical flow estimation, 
targeted at large displacements with significant occlusions. It 
consists of two steps: 
i) dense matching by edge-preserving interpolation from a sparse set
of matches;  ii)~variational energy minimization initialized with the
dense matches. 
The sparse-to-dense interpolation relies on an appropriate choice of the
distance, namely an edge-aware geodesic distance. This distance is tailored to
handle occlusions and motion boundaries -- two common and difficult
issues for optical flow computation. We also propose an approximation
scheme for the geodesic distance to allow fast computation
without loss of performance.  
Subsequent to the dense interpolation step, standard one-level
variational energy minimization is carried out on the dense matches to
obtain the final flow estimation.  
The proposed approach, called  Edge-Preserving Interpolation 
of Correspondences \emph{(EpicFlow)} is fast and robust to large
displacements. It significantly outperforms the 
state of the art on MPI-Sintel and performs on par 
on Kitti and Middlebury. 
\end{abstract}

\section{Introduction}
\label{sec:intro}

Accurate estimation of optical flow from real-world videos 
remains a challenging problem~\cite{sintel}, 
despite the abundant literature on the topic. 
The main remaining challenges are occlusions, motion discontinuities and large displacements, all present in 
real-world videos. 

Effective approaches were previously proposed for handling the case of 
small displacements (\ie, less than a few pixels) 
\cite{Horn1981,Werlberger2009,Papenberg2006}.
These approaches cast the optical flow problem into an energy minimization framework, often solved using efficient coarse-to-fine algorithms~\cite{Bro04a,sun2014}. 
However, due to the complexity of the minimization,
such methods get stuck in local minima and may fail to
estimate large displacements, which often occur due to fast motion. 
This problem has recently received significant attention.
State-of-the-art approaches~\cite{Bro11a,mdpof} use descriptor matching between adjacent frames
together with the integration of these matches in a variational approach. 
Indeed, matching operators are robust to large displacements and 
motion discontinuities \cite{Bro11a,DeepFlow}.
Energy minimization is carried out in a coarse-to-fine scheme
in order to obtain a full-scale dense flow field guided by the matches. 
A major drawback of coarse-to-fine schemes is error-propagation,
\ie, errors at coarser levels, where different motion layers can overlap, 
can propagate across scales. Even if coarse-to-fine techniques work well in most cases, 
we are not aware of a theoretical guarantee or proof of convergence.   

\begin{figure}
\centering 
\includegraphics[width=0.49\linewidth]{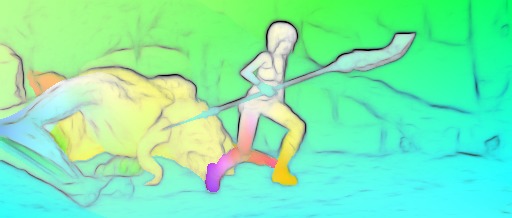}
\hfill
\includegraphics[width=0.49\linewidth]{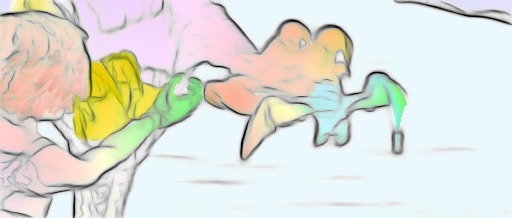}
\caption{Image edges detected with SED~\cite{DollarICCV13edges} and ground-truth
optical flow. Motion discontinuities appear most of the time at image edges.}
\label{fig:contour_boundaries} 
\end{figure}

Instead, we propose to simply interpolate a sparse set of matches in a
dense manner to initiate the optical flow estimation.  We then use
this estimate to initialize a one-level energy minimization,
and obtain the final optical flow estimation.   
This enables us to leverage recent advances in matching algorithms, 
which can now output quasi-dense correspondence fields~\cite{Barnes2010,DeepFlow}.
In the same spirit as~\cite{Leordeanu2013}, we perform a sparse-to-dense interpolation
by fitting a local affine model at each pixel based on nearby matches. 
A major issue arises for the preservation of motion boundaries.
We make the following observation: \emph{motion boundaries often
  tend to appear at image edges}, see Figure~\ref{fig:contour_boundaries}.
Consequently, we propose to exchange the Euclidean distance with a
better, \ie, edge-aware, distance 
and show that this offers a natural way to handle motion discontinuities.
Moreover, we show how an approximation of the edge-aware distance allows
to fit only one affine model per input match (instead of one per pixel). 
This leads to an important speed-up of the interpolation scheme without
loss in performance. 

\begin{figure}[htbp]
\centering 
\includegraphics[width=0.99\linewidth,bb=55bp 100bp 775bp 570bp,clip]{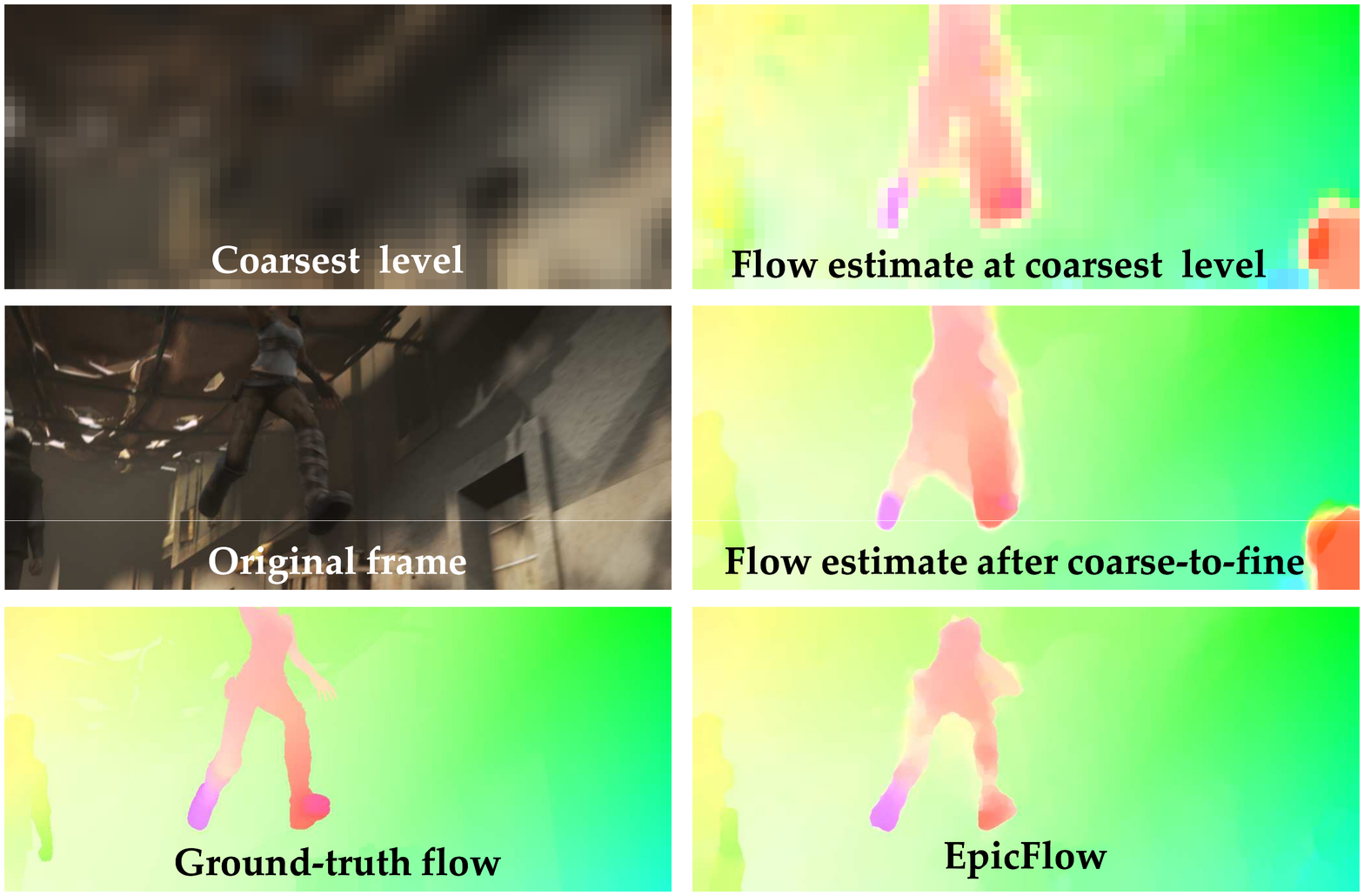}
\caption{Comparison of coarse-to-fine flow estimation and EpicFlow.
Errors at the coarsest level of estimation, due to a low
resolution, often get propagated to the finest level (right, top and
middle).
In contrast, our interpolation scheme benefits from an edge
prior at the finest level (right, bottom).} 
\label{fig:c2f} 
\end{figure}

\begin{figure*}[t]
\centering 
\includegraphics[width=\textwidth]{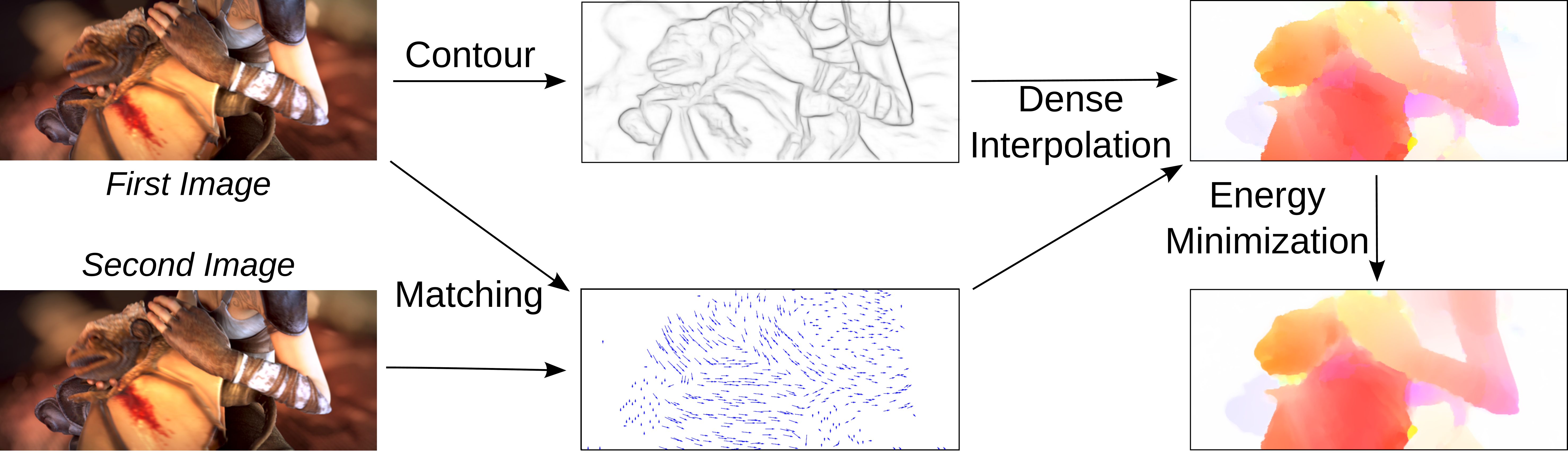}
\caption{Overview of EpicFlow. Given two images, we compute matches
using DeepMatching~\cite{DeepFlow} and the edges of the first image using SED~\cite{DollarICCV13edges}. 
We combine these two cues to densely interpolate matches and obtain a
dense correspondence field. This is used as initialization
of a one-level energy minimization framework.}
\label{fig:densification} 
\end{figure*}

The obtained interpolated field of correspondences is sufficiently
accurate to be used as initialization of a one-level energy minimization. 
Our work suggests that there may be better initialization strategies 
than the well-established coarse-to-fine scheme, see Figure~\ref{fig:c2f}.
In particular, our approach, \emph{EpicFlow} (edge-preserving
interpolation of correspondences) performs best on the challenging
MPI-Sintel dataset~\cite{sintel} and is competitive on
Kitti~\cite{kitti} and Middlebury~\cite{middlebury}. 
An overview of EpicFlow is given in Figure~\ref{fig:densification}. To
summarize, we make three main contributions: 

\noindent $\bullet$~We propose \emph{EpicFlow}, a novel sparse-to-dense interpolation
scheme of matches based on an edge-aware distance. We show that it is robust
to motion boundaries, occlusions and large displacements. \newline
\noindent $\bullet$~We propose an approximation scheme for the edge-aware distance, 
leading to a significant speed-up without loss of accuracy.\newline
\noindent $\bullet$~We show empirically that the proposed optical flow estimation scheme is more accurate than estimations based on coarse-to-fine minimization.

This paper is organized as follows. In Section~\ref{sec:related},
we review related work on large displacement optical flow. 
We then present the sparse-to-dense interpolation in Section~\ref{sec:epic} 
and the energy minimization for optical flow computation in Section~\ref{sec:refine}.
Finally, Section~\ref{sec:experiments} presents  experimental results.
Source code is available online at \url{http://lear.inrialpes.fr/software}.

\section{Related Work}
\label{sec:related}

Most optical flow approaches are based on a variational formulation 
and a related energy minimization problem~\cite{Horn1981,middlebury,sun2014}.
The minimization is carried out using a coarse-to-fine scheme~\cite{Bro04a}.
While such schemes are attractive from a computational point of view, the minimization often gets stuck in local minima 
and leads to error accumulation across scales, especially in the case of large displacements~\cite{alvarezLDOF,Bro11a}.

To tackle this issue, the addition of descriptor/matching was recently investigated in several papers. 
A penalization of the difference between flow and HOG matches was added to the energy by Brox and Malik~\cite{Bro11a}.
Weinzaepfel~\etal~\cite{DeepFlow} 
replaced the HOG matches by an approach based on similarities of non-rigid patches: DeepMatching.
Xu \etal~\cite{mdpof} merged the estimated flow with matching candidates 
at each level of the coarse-to-fine scheme.
Braux-Zin \etal~\cite{Braux-Zin_2013_ICCV} used segment features
in addition to keypoints. 
However, these methods rely on a coarse-to-fine scheme, 
that suffers from intrinsic flaws. Namely,
details are lost at coarse scales, 
and thin objects with substantially different motions cannot be detected.
Those errors correspond to local minima, hence they cannot be recovered and 
are propagated across levels, see Figure~\ref{fig:c2f}.

In contrast, our approach is conceptually closer to recent work that
rely mainly on descriptor
matching~\cite{WillsAB03,PMFilter2013,conf/cvpr/ChenJLCW13,Leordeanu2013,Wills_afeature-based,5459364,bao2014tipeppm}.  
Lu \etal~\cite{PMFilter2013} propose a variant of PatchMatch~\cite{Barnes2010}, which uses SLIC superpixels~\cite{SLIC} as basic blocks in order to better respect
image boundaries. The purpose is to produce a nearest-neighbor-field (NNF) which is later translated into a flow.
However, SLIC superpixels are only \emph{locally} aware of image edges, whereas our edge-aware distance is able to capture regions
at the image scale.
Similarly, Chen \etal~\cite{conf/cvpr/ChenJLCW13} propose to compute an approximate NNF, 
and then estimate the dominant motion patterns using RANSAC.
They, then, use a multi-label graph-cut to solve the assignment of each pixel to a motion pattern candidate. 
Their multi-label optimization can be interpreted as a motion segmentation problem or as a layered model~\cite{DBLP:conf/nips/SunSB10}. 
These problems are hard and a small error in the assignment can lead to large errors in the resulting flow. 

In the same spirit as our approach, Ren~\cite{Ren2008} proposes to use edge-based affinities to group pixels and estimate a piece-wise affine flow.
Nevertheless, this work relies on a discretization of the optical flow constraint, which is valid only for small displacements.
Closely related to EpicFlow, Leordeanu \etal~\cite{Leordeanu2013} also investigate sparse-to-dense interpolation. 
Their initial matching is obtained through the costly minimization of a global non-convex matching energy.
In contrast, we directly use state-of-the-art matches~\cite{DeepFlow,he2012computing} as input.
Furthermore, during their sparse-to-dense interpolation, they compute an affine transformation independently for each pixel based on its neighborhood matches, 
which are found in a Euclidean ball and weighted by an estimation of occluded areas that involves learning a binary classifier. 
In contrast, we propose to use an edge-preserving distance that naturally handles occlusions, and can be very efficiently computed.

\section{Sparse-to-dense interpolation}
\label{sec:epic}

The proposed approach, EpicFlow, consists of three steps, as illustrated in Figure~\ref{fig:densification}.
First, we compute a sparse set of matches between the two images, 
using a state-of-the-art matching algorithm. Second, we perform a densification of this set of matches, 
by computing a sparse-to-dense interpolation from the sparse set of matches, which yields
an initial estimate of the optical flow. Third, we compute the final optical flow estimation by performing
one step of variational energy minimization using the dense interpolation as initialization, see Section~\ref{sec:refine}.

\subsection{Sparse set of matches}

The first step of our approach extracts a sparse set of matches, see  Figure~\ref{fig:densification}. 
Any state-of-the-art matching algorithm can be used to compute the initial set of sparse matches.
In our experiments, we compare the results when using DeepMatching~\cite{DeepFlow} or a subset of an estimated nearest-neighbor field~\cite{he2012computing}. 
We defer to Section~\ref{sub:xpmatch} for a description of these matching algorithms. 
In both cases, we obtain $\sim 5000$ matches for an image of resolution $1024\times436$, \ie, an average of around one match per 90 pixels.
We also evaluate the impact of matching quality and density on the performance of EpicFlow by generating artificial matches 
from the ground-truth in Section~\ref{sec:epicflow-vs-ctf-exps}.
In the following, we denote by $\mathcal{M} = \{ (\bm{p}_m, \bm{p'}_m) \}$ the sparse set of input matches, where each match 
$(\bm{p}_m, \bm{p'}_m)$ defines a correspondence between a pixel $\bm{p}_m$ in the first image and
and a pixel $\bm{p}'_m$ in the second image.
 
\subsection{Interpolation method}
\label{sub:dense-ops}

We estimate a dense correspondence field $\bm{F}: I \rightarrow I'$ between a source image $I$ and a target image $I'$ by interpolating
a sparse set of inputs matches $\mathcal{M} = \{ (\bm{p}_m, \bm{p'}_m) \}$.
The interpolation requires a distance $D:I\times I \rightarrow\mathbb{R}^{+}$ between pixels, see Section~\ref{sub:geo_dist}. 
We consider here two options for the interpolation.

\noindent $\bullet$~\textbf{Nadaraya-Watson (NW) estimation~\cite{Wasserman2010}.}
The correspondence field $\bm{F}_{NW}(\bm{p})$ is interpolated using the Nadaraya-Watson estimator 
at a pixel $\bm{p} \in I$ and is expressed by a sum of matches weighted by their proximity to $\bm{p}$: 
\begin{equation}
\bm{F}_{NW}(\bm{p})= \frac{ \underset{(\bm{p}_m,\bm{p'}_m) \in \mathcal{M} }{\sum} k_{D}(\bm{p}_{m},\bm{p})\bm{p'}_{m} }
		{ \underset{(\bm{p}_m,\bm{p'}_m) \in \mathcal{M} }{\sum} k_{D}(\bm{p}_{m},\bm{p})} ~~,
\label{eqn:nadaraya}
\end{equation}
where $k_{D}(\bm{p}_{m},\bm{p})=\exp\left(-aD(\bm{p}_{m},\bm{p})\right)$ is a Gaussian kernel for a distance $D$ with a parameter $a$.

\noindent $\bullet$~\textbf{Locally-weighted affine (LA) estimation~\cite{Hartley:2003:MVG:861369}.}
The second estimator is based on fitting a local affine transformation.
The correspondence field $\bm{F}_{LA}(\bm{p})$ is interpolated using a
locally-weighted affine estimator at a pixel $\bm{p} \in I$ as  
$\bm{F}_{LA}(\bm{p})=A_{\bm{p}} \bm{p}+t^{\top}_{\bm{p}}$~,
where $A_{\bm{p}}$ and $t_{\bm{p}}$ are the parameters of an affine transformation estimated for pixel $\bm{p}$.
These parameters are computed as the least-square solution of an overdetermined system obtained by writing two equations 
for each match $(\bm{p}_m,\bm{p'}_m) \in \mathcal{M}$ weighted as previously:
\begin{equation}
k_{D}(\bm{p}_{m},\bm{p})\left(A_{\bm{p}} \bm{p}_m+t^{\top}_{\bm{p}}-\bm{p'}_{m}\right)=0 ~~.
\label{eqn:affine1match}
\end{equation}

\noindent \textbf{Local interpolation.}
Note that the influence of remote matches is either negligible, or could harm
the interpolation, for example when objects move differently. Therefore, we restrict the set of matches
used in the interpolation at a pixel $\bm{p}$ to its $K$ nearest
neighbors according to the distance $D$, which we denote as $\mathcal{N}_{K}(\bm{p})$.
In other words, we replace the summation over $\mathcal{M}$ in the
NW operator by a summation over $\mathcal{N}_{K}(\bm{p})$,
and likewise for building the overdetermined system to fit the affine transformation for $\bm{F}_{LA}$.

\subsection{Edge-preserving distance}
\label{sub:geo_dist}

\begin{figure*}
\centering 
\begin{tabular}{c@{ }c@{ }c@{ }c}
\includegraphics[width=0.245\linewidth]{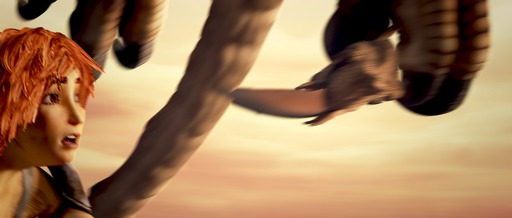} & 
\includegraphics[width=0.245\linewidth]{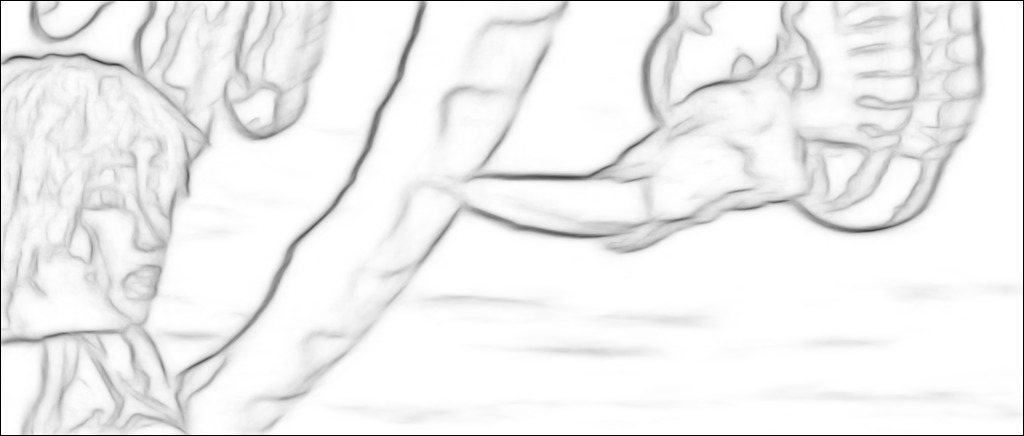} & 
\includegraphics[width=0.245\linewidth]{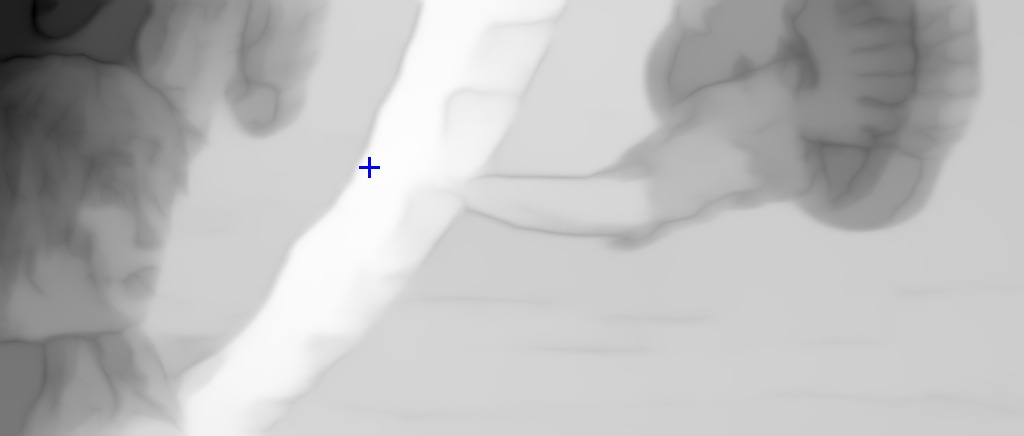} &
\includegraphics[width=0.245\linewidth]{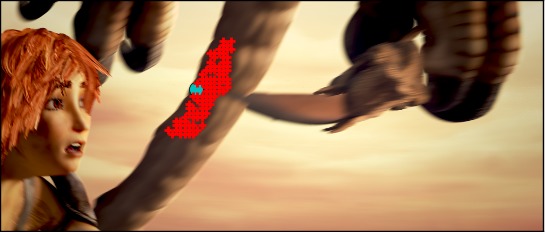} \\[-0.25cm]
\tiny (a) & \tiny (c) & \tiny (e) & \tiny (g) \normalsize \\[-0.1cm]
\includegraphics[width=0.245\linewidth]{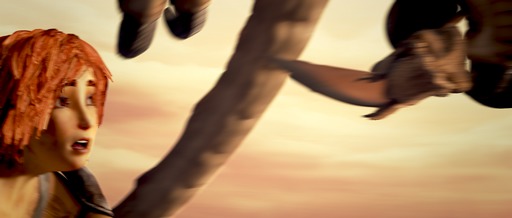} & 
\includegraphics[width=0.245\linewidth]{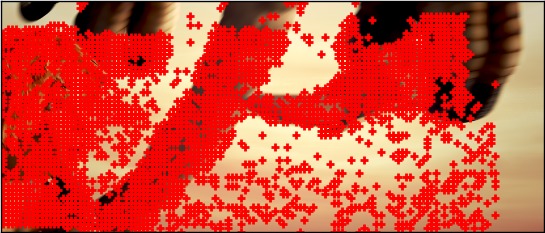} & 
\includegraphics[width=0.245\linewidth]{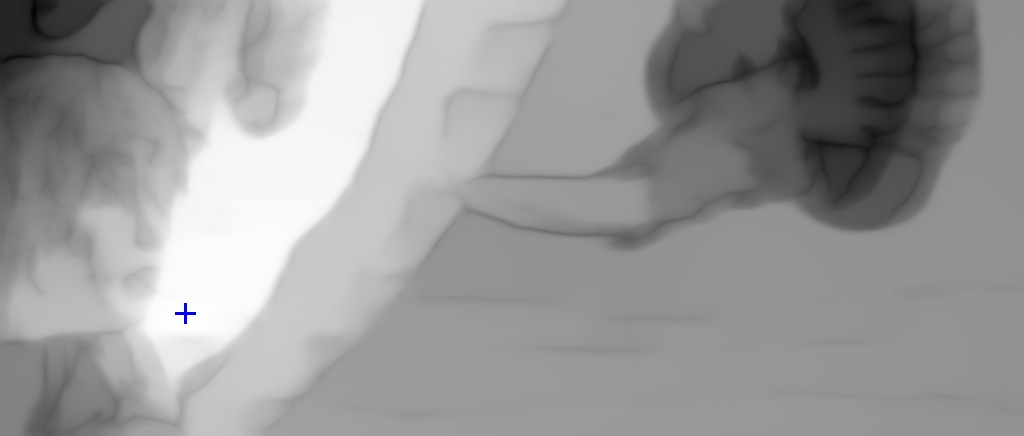} &
\includegraphics[width=0.245\linewidth]{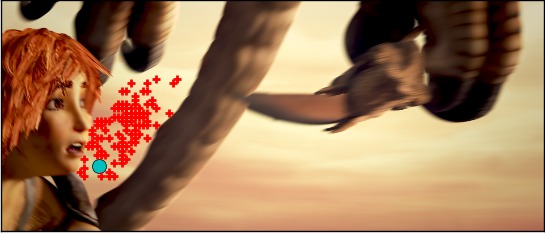} \\[-0.25cm]
\tiny (b) & \tiny (d) & \tiny (f) & \tiny (h) \normalsize
\end{tabular}
\caption{ (a-b) two consecutive frames; (c) 
contour response $C$ from SED~\cite{DollarICCV13edges} (the darker, the higher); (d) match positions $\{ \bm{p}_m \}$ from DeepMatching~\cite{DeepFlow}; (e-f) geodesic
distance from a pixel $\bm{p}$ (marked in blue) to all others $D_G(\bm{p},.)$ (the brighter, the closer). (g-h)
100 nearest matches, \ie, $\mathcal{N}_{100}(\bm{p})$ (red) using geodesic distance $D_G$ from the pixel $\bm{p}$ in blue.}
\label{fig:sed} 
\end{figure*}

Using the Euclidean distance for the interpolation presented above is
possible. However, in this case the interpolation is simply
based on the position of the input matches and does not 
respect motion boundaries.  
Suppose for a moment that the motion boundaries are known.
We can, then, use a geodesic distance $D_G$ based on these motion boundaries.
Formally, the geodesic distance between two pixels $\bm{p}$ and $\bm{q}$ 
is defined as the shortest distance with respect to a cost map $C$: 
\begin{equation}
D_G(\bm{p},\bm{q})=\inf_{\Gamma\in\mathcal{P}_{\bm{p},\bm{q}}}\int_{\Gamma}C(\bm{p}_{s})d\bm{p}_{s}~~,\label{eq:dt}
\end{equation}
where $\mathcal{P}_{\bm{p},\bm{q}}$ denotes the set of all possible paths between
$\bm{p}$ and $\bm{q}$, and $C(\bm{p}_{s})$ the cost of crossing pixel $\bm{p}_{s}$ (the viscosity in physics).
In our settings, $C$ corresponds to the motion boundaries.
Hence, a pixel belonging to a motion layer is close to all other pixels from the same
layer according to $D_G$, but far from everything beyond the boundaries. Since each pixel is interpolated
based on its neighbors, the interpolation will respect the motion boundaries.

In practice, we use an alternative to true motion boundaries, making the 
plausible assumption that \emph{image edges} are a superset of \emph{motion
boundaries}.
This way, the distance between pixels belonging to the same region will be low.
It ensures a proper edge-respecting interpolation as long as the number of 
matches in each region is sufficient.
Similarly, Criminisi \etal~\cite{Criminisi2010} showed that geodesic distances
are a natural tool for edge-preserving image editing operations (denoising, texture flattening, etc.) 
and it was also used recently to generate object proposals~\cite{geodesicproposal}.
In practice, we set the cost map $C$ using a recent state-of-the-art edge detector, namely the ``structured
edge detector''
(SED)~\cite{DollarICCV13edges}\footnote{https://github.com/pdollar/edges}. 
Figure~\ref{fig:sed} shows an example of a SED map, 
as well as examples of geodesic distances and neighbor sets $\mathcal{N}_{K}(\bm{p})$ for different pixels $\bm{p}$.
Notice how neighbors are found on the same objects/parts of the image with $D_G$, 
in contrast to Euclidean distance (see also Figure~\ref{fig:occluded}).

\subsection{Fast approximation}
\label{sub:epic-approx}

The geodesic distance can be rapidly computed from a point to all other pixels.
For instance, Weber \etal~\cite{Weber2008_DT} propose parallel algorithms that simulate an advancing wavefront.
Nevertheless, the computational cost for computing the geodesic distance between all pixels and all matches (as required by our interpolation scheme) is high. 
We now propose an efficient approximation $\tilde{D}_G$.

A key observation is that neighboring pixels are often 
interpolated similarly, suggesting a strategy that would leverage  such local information. 
In this section we employ the term `match' to refer to $\bm{p}_m$ instead of $(\bm{p}_m,\bm{p'}_m)$.

\begin{figure}
\centering 
\includegraphics[width=\linewidth]{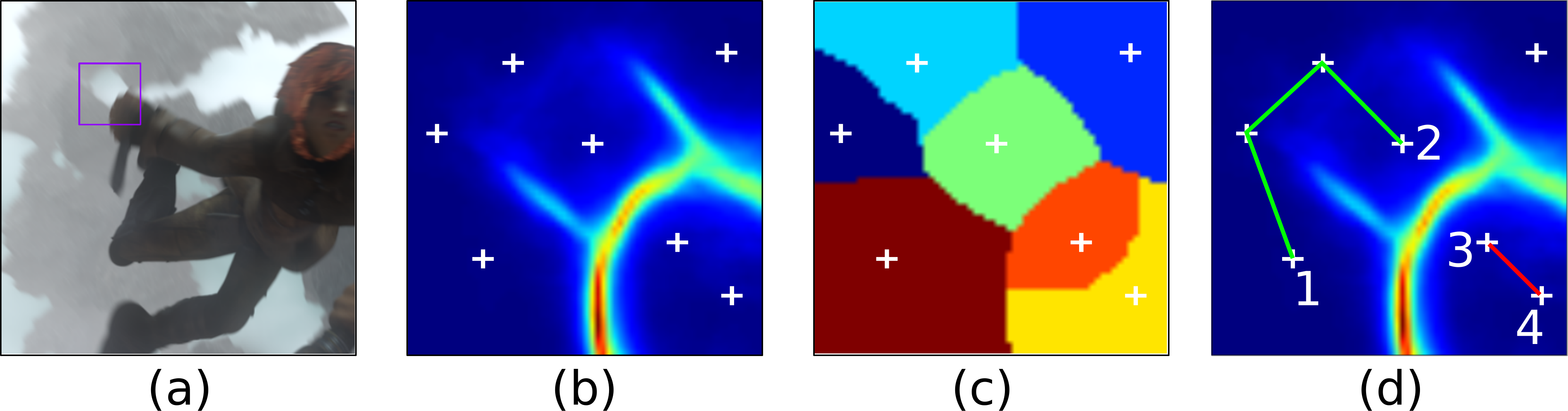} 
\caption{For the region shown in (a), (b) shows the image edges $C$ and white crosses representing the match positions $\{ \bm{p}_m \}$. 
(c) displays the assignment $L$, \ie, geodesic Voronoi cells. We build a graph
$\mathcal{G}$ from $L$ (see text). (d) shows the shortest path between two neighbor matches, which can go through the edge that connects them (3-4) or 
a shorter path found by Dijkstra's algorithm (1-2).}
\label{fig:voronoi} 
\end{figure}

\noindent \textbf{Geodesic Voronoi diagram.}
We first define a clustering $L$, such that $L(\bm{p})$ assigns a pixel $\bm{p}$
to its closest match according to the geodesic distance, \ie, we
have $L(\bm{p})=\argmin_{\bm{p}_{m}}D_G(\bm{p},\bm{p}_{m})$.
$L$ defines geodesic Voronoi cells, as shown in Figure~\ref{fig:voronoi}(c).

\noindent \textbf{Approximated geodesic distance.}
We then approximate the distance between a pixel $\bm{p}$ and any match
$\bm{p}_{m}$ as the distance to the closest match $L(\bm{p})$ plus an approximate
distance between matches:
\begin{equation}
\tilde{D}_G(\bm{p},\bm{p}_{m})=D_G(\bm{p},L(\bm{p}))+D^{\mathcal{G}}_G(L(\bm{p}),\bm{p}_{m})
\label{eqn:approxdis}
\end{equation}
where $D_G^{\mathcal{G}}$ is a graph-based approximation of the geodesic distance
between two matches. To define $D_G^{\mathcal{G}}$ we use
a neighborhood graph $\mathcal{G}$ whose nodes are $\{\bm{p}_{m}\}$. Two
matches $\bm{p}_{m}$ and $\bm{p}_{n}$ are connected by an edge if they are neighbors
in $L$. The edge weight is then defined as the geodesic
distance between $\bm{p}_m$ and $\bm{p}_n$, where the geodesic distance calculation is restricted to the  
Voronoi cells of $\bm{p}_m$ and $\bm{p}_n$. We, then, calculate the
approximate geodesic distance between any two matches $\bm{p}_{m},\bm{p}_{n}$ 
using Dijkstra's algorithm on $\mathcal{G}$, see Figure~\ref{fig:voronoi}(d).

\noindent \textbf{Piecewise field.}
So far, we have built an approximation of the distance between
pixels and match points. We now show that our interpolation
model results in a piece-wise correspondence field (either constant for the Nadaraya-Watson
estimator, or piece-wise affine for LA). This property is crucial to obtain
a fast interpolation scheme, and experiments shows that it
does not impact the accuracy. Let us consider a pixel $\bm{p}$ such that $L(\bm{p}) = \bm{p}_m$.
The distance between $\bm{p}$ and any match
$\bm{p}_n$ is  the same as the one between $\bm{p}_m$ and $\bm{p}_n$ up to 
a constant independent from $\bm{p}_n$ (Equation~\ref{eqn:approxdis}).
As a consequence, we have $\mathcal{N}_{K}(\bm{p})=\mathcal{N}_{K}(\bm{p}_{m})$ and $k_{\tilde{D}_G}(\bm{p},\bm{p}_n) = k_{D_G}(\bm{p},\bm{p}_m) \times k_{D_G^{\mathcal{G}}}(\bm{p}_{m},\bm{p}_{n})$.
For the Nadaraya-Watson estimator, we thus obtain:
\begin{eqnarray}
& \bm{F}_{NW}(\bm{p}) = \frac{ \sum_{(\bm{p}_n,\bm{p'}_n)} k_{\tilde{D}_G}(\bm{p},\bm{p}_n) \bm{p'}_{n}  }{ \sum_{(\bm{p}_n,\bm{p'}_n)} k_{\tilde{D_G}}(\bm{p},\bm{p}_n) } \\
& = \frac{ k_{D_G}(\bm{p},\bm{p}_m) \sum_{(\bm{p}_n,\bm{p'}_n)} k_{D_G^{\mathcal{G}}}(\bm{p}_{m},\bm{p}_{n}) \bm{p'}_{n} }{ k_{D_G}(\bm{p},\bm{p}_m) \sum_{(\bm{p}_n,\bm{p'}_n)} k_{D_G^{\mathcal{G}}}(\bm{p}_{m},\bm{p}_{n}) } = \bm{F}_{NW}(\bm{p}_m) \nonumber
\end{eqnarray}
where all the sums are for $(\bm{p}_n,\bm{p'}_n) \in \mathcal{N}_{K}(\bm{p})=\mathcal{N}_{K}(\bm{p}_m)$.
 The same reasoning holds for the weighted affine interpolator, which
is invariant to a multiplication of the weights by a constant factor.
As a consequence, it suffices to compute $\left|\mathcal{M}\right|$
estimations (one per match) and to propagate it to the pixel
assigned to this match. This is orders of magnitude faster than an
independent estimation for each
pixel, \eg as done in~\cite{Leordeanu2013}. We summarize
the approach in Algorithm~\ref{alg:wmm} for Nadaraya-Watson estimator. The algorithm is similar 
for LA interpolator (\eg line 6 becomes 
``Estimate affine parameters $A_{\bm{p}_m}, t_{\bm{p}_m}$'' 
and line 8 ``Set $\bm{W}_{LA}(\bm{p}) = A_{L(\bm{p})} \bm{p} + t^{\top}_{L(\bm{p})}$'').

\begin{algorithm}
\small
\textbf{Input:} a pair of images $I,I'$, a set $\mathcal{M}$ of matches

\textbf{Output:} dense correspondence field $\bm{F}_{NW}$

1 $\quad$ Compute the cost $C$ for $I$ using SED~\cite{DollarICCV13edges}

2 $\quad$ Compute the assignment map $L$

3 $\quad$ Build the graph $\mathcal{G}$ from $L$

4 $\quad$ \textbf{For} $(\bm{p}_m,\bm{p'}_m) \in\mathcal{M}$

5 $\quad$ $\quad$Compute $\mathcal{N}_{K}(\bm{p}_{m})$ from $\mathcal{G}$
using Dijkstra's algorithm

6 $\quad$ $\quad$Compute $\bm{F}_{NW}(\bm{p}_m)$ from $\mathcal{N}_{K}(\bm{p}_{m})$
using Eq.~\ref{eqn:nadaraya}

7 $\quad$ \textbf{For} each pixel $\bm{p}$

8 $\quad$ $\quad$Set $\bm{F}_{NW}(\bm{p}) = \bm{F}_{NW}(L(\bm{p}))$

\normalsize
\caption{Interpolation with Nadaraya-Watson}
\label{alg:wmm} 
\end{algorithm}

\section{Optical Flow Estimation}
\label{sec:refine}

\noindent \textbf{Coarse-to-fine vs. EpicFlow.}
The output of the sparse-to-dense interpolation is a dense correspondence field.
This field is used as initialization of a variational energy minimization method. 
In contrast to our approach, state-of-the-art methods usually rely on a coarse-to-fine scheme to compute the full-scale correspondence field. 
To the best of our knowledge, there exists no theoretical proof or guarantee that a coarse-to-fine minimization 
leads to a consistent estimation that accurately minimizes the full-scale energy. 
Thus, the coarse-to-fine scheme should be considered as a heuristic 
to provide an initialization for the full-scale flow. 

Our approach can be thought of as an alternative to the above strategy, 
by offering a smart heuristic to accurately initialize
the optical flow before performing energy minimization at the full-scale. 
This offers several advantages over the coarse-to-fine scheme. 
First, the cost map $C$ in our method acts as a prior on boundary location.
Such a prior could also be incorporated by a local smoothness weight in the coarse-to-fine minimization, but would 
then be difficult to interpret at coarse scales where boundaries might strongly overlap.
In addition, since our method directly works at the full image resolution, it avoids
possible issues related to the presence of thin objects that could be oversmoothed at coarse scales. 
Such errors at coarse scales are propagated to finer scales as the coarse-to-fine approach proceeds, see Figure~\ref{fig:c2f}. 

\noindent \textbf{Variational Energy Minimization.}
We minimize an energy defined as a sum of a data term and a smoothness term.
We use the same data term as~\cite{Zimmer2011}, based on a classical color-constancy and gradient-constancy assumption
with a normalization factor.
For the smoothness term, we penalize the flow gradient norm, with a local smoothness weight $\alpha$ as in~\cite{5459375,mdpof}:
$ \alpha(\bm{x}) = \exp \big( - \kappa \Vert \nabla_2 I (\bm{x})\Vert \big)$
with $\kappa=5$. We have also experimented using SED instead and obtained similar performance.

For minimization, we initialize the solution with the output of our sparse-to-dense interpolation and use the approach of~\cite{Bro04a} without the coarse-to-fine scheme.
More precisely, we perform 5 fixed point iterations, \ie, compute the non-linear weights (that appear when applying Euler-Lagrange equations~\cite{Bro04a}) and
the flow updates 5 times iteratively. The flow updates are computed by solving linear systems using 30 iterations of the successive over relaxation method~\cite{sor}.

\section{Experiments}
\label{sec:experiments}

In this section, we evaluate EpicFlow on three state-of-the-art datasets:

\noindent $\bullet$~\emph{MPI-Sintel dataset}~\cite{sintel}
is a challenging evaluation benchmark obtained from an
animated movie. It contains multiple sequences including large/rapid motions. 
We only use the `final' version that features realistic rendering effects such as
motion, defocus blur and atmospheric effects. \newline
\noindent $\bullet$~The \emph{Kitti dataset}~\cite{kitti} contains
photos shot in city streets from a driving platform. It features large
displacements, different materials (complex 3D objects like trees),
a large variety of lighting conditions and non-lambertian surfaces.
\noindent $\bullet$~The \emph{Middlebury dataset}~\cite{middlebury}
has been extensively used for evaluating optical flow methods.
It contains complex motions, but displacements are limited to a few pixels.

As in~\cite{DeepFlow}, we optimize the parameters on a subset (20\%) of the MPI-Sintel training set. 
We then report average endpoint error (AEE) on the remaining MPI-Sintel training set (80\%), the Kitti training set and the Middlebury training set.
This allows us to evaluate the impact of parameters on different datasets and avoid overfitting.
The parameters are typically $a \simeq 1$ for the coefficient in the kernel $k_D$, the number of neighbors is $K \simeq 25$ for NW interpolation and $K \simeq 100$
when using LA.
In Section~\ref{sec:xptest}, we compare to the state of the art on the test sets.
In this case, the parameters are optimized on the training set of the corresponding dataset.
Timing is reported for one CPU-core at 3.6GHz.

In the following, we first describe two types of input matches in Section~\ref{sub:xpmatch}.
Section~\ref{sec:xpfeatures} then studies the different parameters of our approach.
In  Section~\ref{sec:epicflow-vs-ctf-exps}, we compare our method to a variational approach with a coarse-to-fine scheme.
Finally, we show that EpicFlow outperforms current methods on challenging datasets in Section~\ref{sec:xptest}.

\subsection{Input matches}
\label{sub:xpmatch}

To generate input matches, we use and compare two recent matching algorithms. They each produce about 5000 matches per image.

\begin{figure}
\centering 
\hfill
\includegraphics[width=0.49\linewidth]{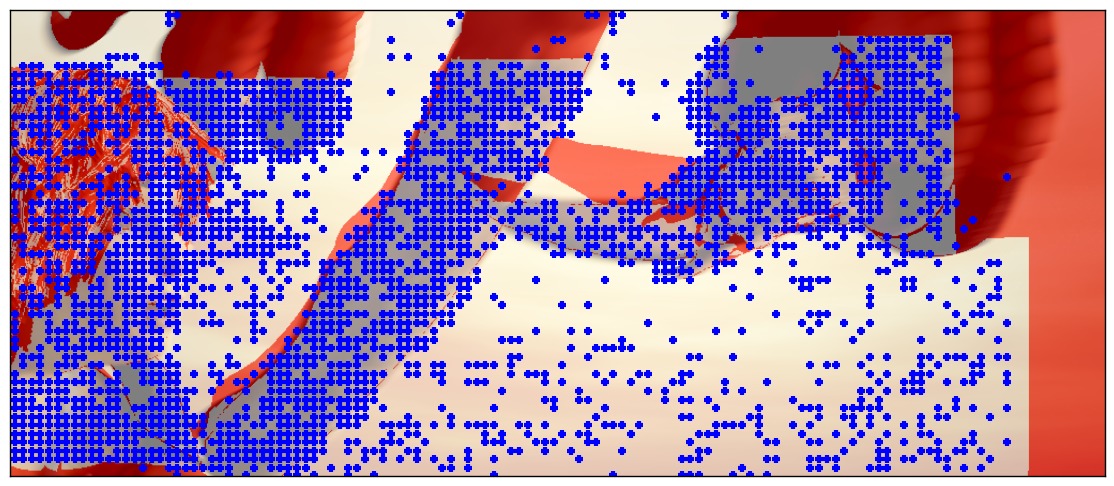}
\hfill
\includegraphics[width=0.49\linewidth]{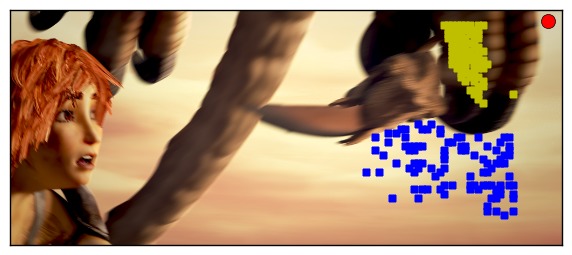}
\hfill
\caption{\emph{Left:} Match positions returned by~\cite{DeepFlow} are shown in blue. Red denotes occluded areas.
\emph{Right:} Yellow (resp. blue) squares correspond to the 100 nearest matches
with a Euclidean (resp. edge-aware geodesic) distance for the occluded pixel shown in red. 
All neighbor matches with a Euclidean distance holds to a different object 
while the geodesic distance allows to capture matches from the same region, even in the case of large occluded areas.}
\label{fig:occluded} 
\end{figure}

\noindent $\bullet$ The first one is DeepMatching (\textbf{DM}), used in DeepFlow~\cite{DeepFlow}, which has shown excellent performance for optical flow.
It builds correspondences by computing similarities of non-rigid patches, allowing for some deformations. 
We use the online code\footnote{http://lear.inrialpes.fr/src/deepmatching/} on images downscaled by a factor~2.
A reciprocal verification is included in DM. As a consequence, the majority of matches in occluded areas are pruned, see 
matches in Figure~\ref{fig:occluded} (left).

\noindent $\bullet$ The second one is a recent variant of PatchMatch~\cite{Barnes2010} that relies on kd-trees and 
local propagation to compute a dense correspondence field~\cite{he2012computing} (\textbf{KPM}). 
We use the online code to extract the dense correspondence field\footnote{http://j0sh.github.io/thesis/kdtree/}. 
It is noisy, as it is based on small patches without global regularization, as well as often incorrect in case of occlusion.  
Thus, we  perform a two-way matching and eliminate non-reciprocal matches to remove incorrect correspondences.  
We also subsample these pruned correspondences to speed-up the interpolation.
We have experimentally verified on several image pairs that this subsampling does not result in a loss of performance.  

\noindent \textbf{Pruning of matches.}
\label{epic:filtering} In both cases, matches are extracted locally and
might be incorrect in regions with low texture. Thus, we remove 
matches corresponding to patches with low saliency, which are
determined by the eigenvalues of autocorrelation matrix.  
Furthermore, we perform a consistency check to remove outliers. 
We run the sparse-to-dense interpolation once with the Nadaraya-Watson estimator
and remove matches for which the difference to the initial estimate is over 5 pixels.

We also experiment with synthetic sparse matches of various densities and noise levels in Section~\ref{sec:epicflow-vs-ctf-exps},
in order to evaluate the sensitivity of EpicFlow to the quality of the matching approach.

\subsection{Impact of the different parameters}
\label{sec:xpfeatures}

In this section, we evaluate the impact of the matches and the
interpolator. We also compare the quality of the sparse-to-dense
interpolation and EpicFlow.  Furthermore, we examime the impact of
the geodesic distance and its approximation as well as the impact of
the quality of the contour detector.

\noindent \textbf{Matches and interpolators.}
Table~\ref{tab:models} compares the result of our sparse-to-dense
interpolation, \ie, before energy
minimization, and EpicFlow for different matches (\textbf{DM} and
\textbf{KPM}) and for the two interpolation schemes: Nadaraya-Watson
(\textbf{NW}) and locally-weighted affine (\textbf{LA}). 
The approximated geodesic distance is used in the interpolation, see Section~\ref{sub:epic-approx}.

We can observe that KPM is consistently outperformed
by DeepMatching (DM) on MPI-Sintel and Kitti datasets, with a gap of 2 and 8 pixels respectively. 
Kitti contains many repetitive textures like trees or roads, which are
often mismatched by KPM. Note that DM is significantly
more robust to repetitive textures than KPM, as it uses a
multi-scale scoring scheme. The results on Middlebury are
comparable and below 1 pixel.

\begin{table}
\centering
\resizebox{\linewidth}{!}{
\begin{tabular}{|c|c|c||c|c|c|}
\hline
& Matching  & Interpolator  & MPI-Sintel & Kitti & Middlebury\\
\hline 
{\multirow{4}{*}{\rotatebox[origin=c]{90}{\footnotesize Interpolation}}} 
&KPM & NW & 6.052 & 15.679 & 0.765 \\
&KPM & LA & 6.334 & 12.011 & 0.776 \\
&DM & NW & 4.143 & 5.460 & 0.898 \\
&DM & LA & 4.068 & 3.560 & 0.840 \\
\hline
{\multirow{4}{*}{\rotatebox[origin=c]{90}{\footnotesize EpicFlow}}}
&KPM & NW & 5.741 & 15.240 & 0.388 \\
&KPM & LA & 5.764 & 11.307 & \textbf{0.315} \\
&DM  & NW & 3.804 & 4.900 & 0.485 \\
&DM  & LA & \textbf{3.686} & \textbf{3.334} & 0.380 \\
\hline
\end{tabular}
\normalsize
}
\caption{Comparison of average endpoint error (AEE) for different
sparse matches (DM, KPM)  and interpolators (NW, LA) as well as for 
sparse-to-dense interpolation (top) and EpicFlow (bottom). The 
approximated geodesic distance $\tilde{D}_G$ is used.}
\label{tab:models}
\end{table}

We also observe that LA performs better than NW on Kitti, while the results are
comparable on MPI-Sintel and Middlebury. This 
is due to the specificity of the Kitti dataset, where the scene
consists of planar surfaces and, thus, affine transformations are more suitable than translations to approximate the flow.
Based on these results, we use DM matches and LA interpolation in the remainder of the experimental section.

The interpolation is robust to the neighborhood size $K$ with for instance an AEE of $4.082, 4.053, 4.068$ and $4.076$ 
for $K=50, 100, 160$ (optimal value on the training set), $200$ respectively, on MPI-Sintel with the LA estimator and before variational minimization.
We also implemented a variant where we use all matches closer than a threshold and obtained similar performance.

\begin{figure*}
\centering
\begin{tabular}{cccc}
\begin{turn}{90}{\small{}{}{}~~~~~~Images}\end{turn}  & \includegraphics[width=0.275\linewidth]{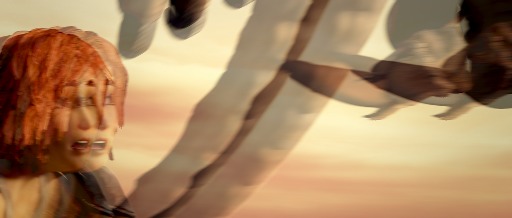}  & \includegraphics[width=0.275\linewidth]{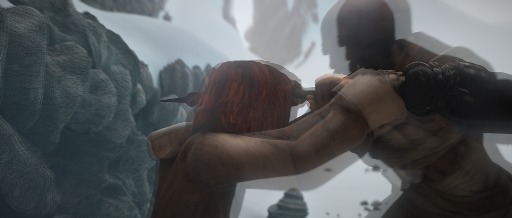}  & \includegraphics[width=0.275\linewidth]{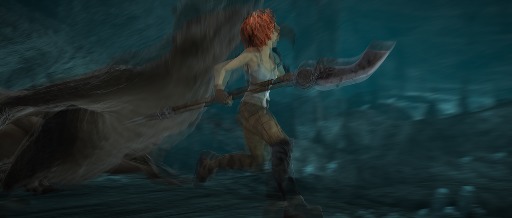} \\[-0.06cm]
\begin{turn}{90}{\small{}{}{}Ground-Truth}\end{turn}  & \includegraphics[width=0.275\linewidth]{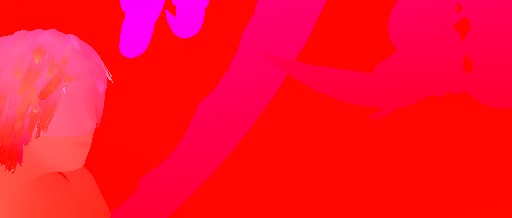}  & \includegraphics[width=0.275\linewidth]{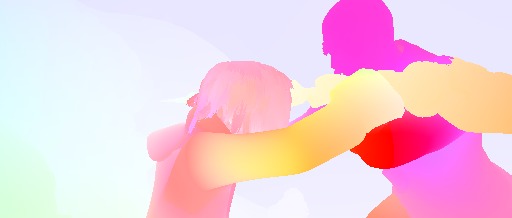}  & \includegraphics[width=0.275\linewidth]{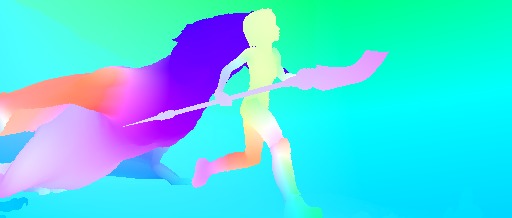} \\[-0.06cm]
\begin{turn}{90}{\small{}{}{}\textbf{Interpolation}}\end{turn}  & \includegraphics[width=0.275\linewidth]{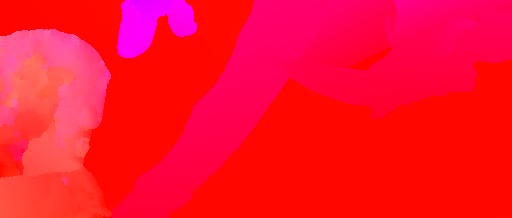}  & \includegraphics[width=0.275\linewidth]{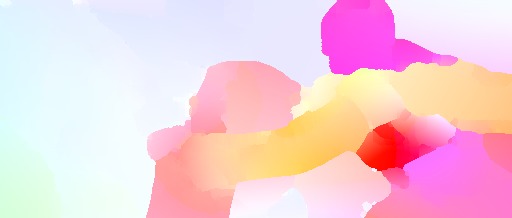}  & \includegraphics[width=0.275\linewidth]{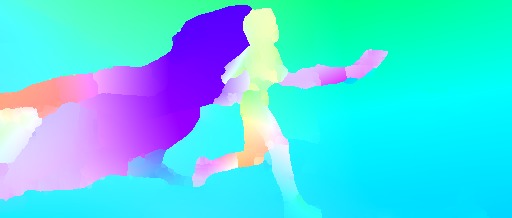} \\[-0.06cm]
\begin{turn}{90}{\small{}{}{}~~~~\textbf{EpicFlow}}\end{turn}  & \includegraphics[width=0.275\linewidth]{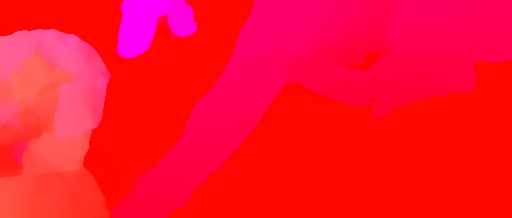}  & \includegraphics[width=0.275\linewidth]{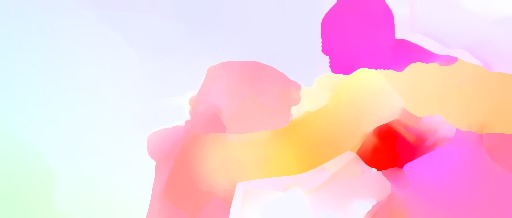}  & \includegraphics[width=0.275\linewidth]{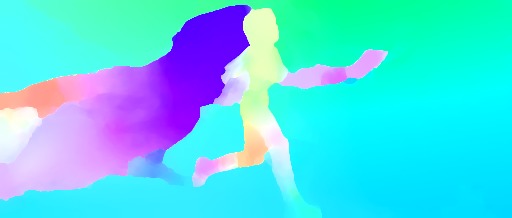} \\[-0.06cm]
\begin{turn}{90}{\small{}{}{}DeepFlow~\cite{DeepFlow}}\end{turn}  & \includegraphics[width=0.275\linewidth]{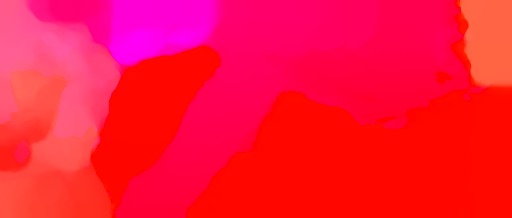}  & \includegraphics[width=0.275\linewidth]{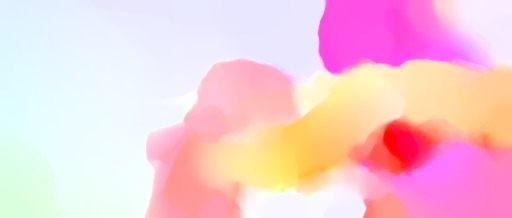}  & \includegraphics[width=0.275\linewidth]{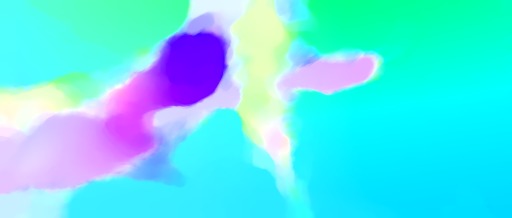} \\[-0.06cm]
\begin{turn}{90}{\footnotesize{}{}{}MDPFlow2~\cite{mdpof}}\end{turn}  & \includegraphics[width=0.275\linewidth]{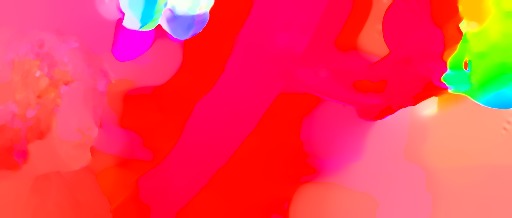}  & \includegraphics[width=0.275\linewidth]{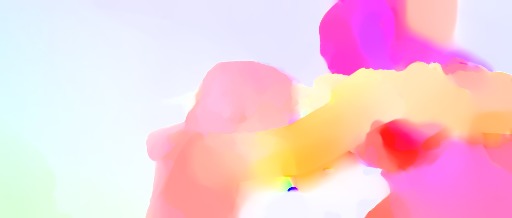}  & \includegraphics[width=0.275\linewidth]{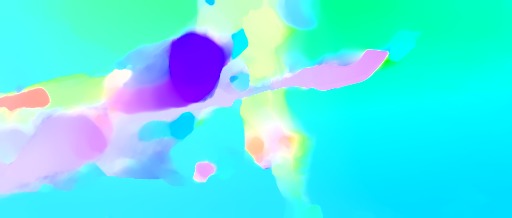} \\[-0.06cm]
\begin{turn}{90}{\small{}{}{}~~~~~LDOF~\cite{Bro11a}}\end{turn}  & \includegraphics[width=0.275\linewidth]{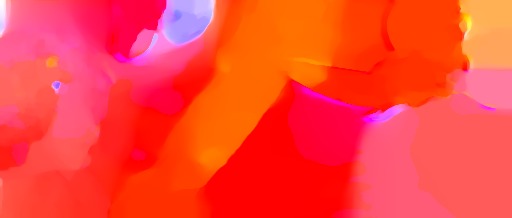}  & \includegraphics[width=0.275\linewidth]{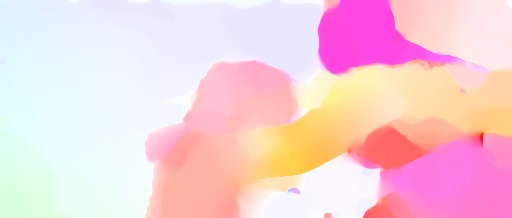}  & \includegraphics[width=0.275\linewidth]{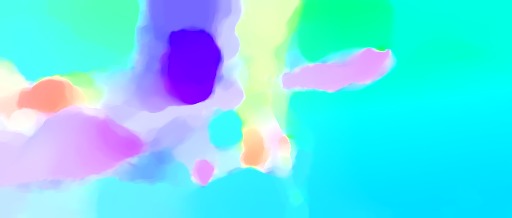}
\normalsize
\end{tabular}
\normalsize
\caption{Each column shows from top to bottom: mean of
two consecutive images, ground-truth flow, result of sparse-to-dense interpolation (Interpolation), full method (EpicFlow), and 3 state-of-the-art methods. 
EpicFlow better respects motion boundaries, is able to capture small parts like the limbs of the character (right column) and successfully estimates the 
flow in occluded areas (right part of left column).}
\label{fig:examples} 
\end{figure*}

\noindent \textbf{Sparse-to-dense interpolation versus EpicFlow.}
We also evaluate the gain due to the variational minimization
using the interpolation as initialization. 
We can see in Table~\ref{tab:models} that this step clearly
improves the performance in all cases. The improvement is around 0.5
pixel. Figure~\ref{fig:examples} presents results for three image
pairs with the initialization only and the final result of EpicFlow
(row three and four). 
While the flow images look similar overall, the minimization allows
to further smooth and refine the flow, explaining the gain in performance.
Yet, it preserves discontinuities and small details, such as the legs
in the right column. In the following, results are reported for
EpicFlow, \ie, after the variational minimization step.

\noindent \textbf{Edge-aware versus Euclidean distances.}
We now study the impact of different distances. First, we examine the
effect of approximating the geodesic distance (Section~\ref{sub:epic-approx}). 
Table~\ref{tab:distance} shows that 
our approximation has a negligible impact when compared to the exact
geodesic distance. Note that the exact version performs distance
computation as well as local estimation per pixel and is, thus, 
an order of magnitude slower to compute, see last column of
Table~\ref{tab:distance}.  

Next, we compare the geodesic distance and Euclidean distances.   
Table~\ref{tab:distance} shows that using a Euclidean distance leads
to a significant drop in performance, in particular for the 
MPI-Sintel dataset, the drop is 1 pixel. This confirms the importance of our
edge-preserving distance. 
Note that the result with the Euclidean distance is reported with an
exact version, \ie, the interpolation is computed pixelwise. 

\begin{table}
\centering
\resizebox{\linewidth}{!}{
\begin{tabular}{|c|l||c|c|c||r|}
\hline
Contour & Distance  & MPI-Sintel  & Kitti & Middlebury & Time \\
\hline 
\textbf{SED}~\cite{DollarICCV13edges} & \textbf{Geodesic (approx.)} & 3.686 & 3.334 & \textbf{0.380} & 16.4s \\
SED~\cite{DollarICCV13edges} & Geodesic (exact) & \textbf{3.677} & \textbf{3.216} & 0.393 & 204s \\
\hline 
-      & Euclidean & 4.617 & 3.663 & 0.442 & 40s \\
SED~\cite{DollarICCV13edges} & mixed  & 3.975 & 3.510 & 0.399 & 300s \\ 
\hline
gPb~\cite{gPb}   & Geodesic (approx.) & 4.161 & 3.437 & 0.430 & 26s \\
Canny~\cite{Canny} & Geodesic (approx.) & 4.551 & 3.308 & 0.488 & 16.4s \\
$\Vert \nabla_2 \mathcal{I} \Vert_2$ & Geodesic (approx.) & 4.061 & 3.399 & 0.388 & 16.4s \\
GT boundaries & Geodesic (approx.)  & 3.588 & & & \\
\hline
\end{tabular}
}
\caption{Comparison of the AEE of EpicFlow (with DM and LA) for
  different distances and different contour extractors. The 
  time (right column) is reported for a MPI-Sintel image pair. } 
\label{tab:distance} 
\end{table}

We also compare to a mixed approach, 
in which the neighbor list $\mathcal{N}_{K}$ is constructed using
the Euclidean distance, but weights $k_{\tilde{D}}(\bm{p}_m,\bm{p})$ are 
set according to the approximate geodesic distance. Table~\ref{tab:distance} shows that
this leads to a drop of performance by around $0.3$ pixels for MPI-Sintel and Kitti. 
Figure~\ref{fig:occluded} illustrates the reason: 
none of the Euclidean neighbor matches (yellow) belong to the region
corresponding to the selected pixel (red), but all of geodesic neighbor
matches (blue) belong to it. This demonstrates
the importance of using an edge-preserving geodesic distance
throughout the whole pipeline, in contrast to~\cite{Leordeanu2013}
who interpolates matches found in a Euclidean neigbhorhood. 

\noindent \textbf{Impact of contour detector.}
We also evaluate the impact of the contour detector in
Table~\ref{tab:distance}, \ie, the SED detector~\cite{DollarICCV13edges} 
is replaced by the Berkeley gPb detector~\cite{gPb} or the Canny edge detector~\cite{Canny}.
Using gPb leads to a small drop in performance (around $0.1$ pixel on
Kitti and $0.5$ on MPI-Sintel) and significantly increases the
computation time. 
Canny edges perform similar to the Euclidean distance. This can be
explained by the insufficient quality of the Canny contours. 
Using the norm of image's gradient improves slightly over gPb. We found that this is due to the presence of holes 
when estimating contours with gPb.
Finally, we perform experiments using ground-truth motion boundaries, computed from the norm of ground-truth flow gradient, and obtain an improvement
of $0.1$ on MPI-Sintel ($0.2$ before the variational part). The ground-truth flow is not dense enough on Middlebury and Kitti datasets to estimate GT boundaries.

\subsection{EpicFlow versus coarse-to-fine scheme}
\label{sec:epicflow-vs-ctf-exps}

To show the benefit of our approach, we have carried out a comparison
with a coarse-to-fine scheme. 
Our implementation of the variational approach is the same as in Section~\ref{sec:refine}, 
with a coarse-to-fine scheme and DeepMatching integrated in the energy
through a penalization of the difference between flow and
matches~\cite{Bro11a,DeepFlow}.  
Table~\ref{tab:epicVSvar} compares EpicFlow to the
variational approach with coarse-to-fine scheme, using exactly the same matches as input. 
EpicFlow performs better and is also faster. 
The gain is around $0.4$ pixel on MPI-Sintel and over $1$ pixel on
Kitti. The important gain on Kitti might be explained by the affine
model used for interpolation, which fits well the piecewise planar
structure of the scene. On Middlebury, the variational approach
achieves slightly better results, as this dataset does not contain
large displacements. 

Figure~\ref{fig:examples} shows a comparison to three state-of-the-art
methods, all built upon a coarse-to-fine scheme.
Note how motion boundaries are preserved by EpicFlow.
Even small details, like the limbs in the right column, are captured.
Similarly, in the case of occluded areas, EpicFlow benefits from the
geodesic distance to produce a correct estimation, see the right part
of the left example. 

\begin{table}
\centering
\resizebox{\linewidth}{!}{
\begin{tabular}{|c||c|c|c||r|}
\hline
Flow method & MPI-Sintel & Kitti & Middlebury & Time \\
\hline 
DM+coarse-to-fine & 4.095 & 4.422 & \textbf{0.
321} & 25s \\
\textbf{DM+EpicFlow} & \textbf{3.686} & \textbf{3.334} & 0.380  & 16.4s \\
\hline
\end{tabular}
}
\caption{Comparison of AEE for EpicFlow (with DM + LA) and a
  coarse-to-fine scheme (with DM).}
\label{tab:epicVSvar}
\end{table}

\begin{figure}
\centering
\includegraphics[width=1\linewidth]{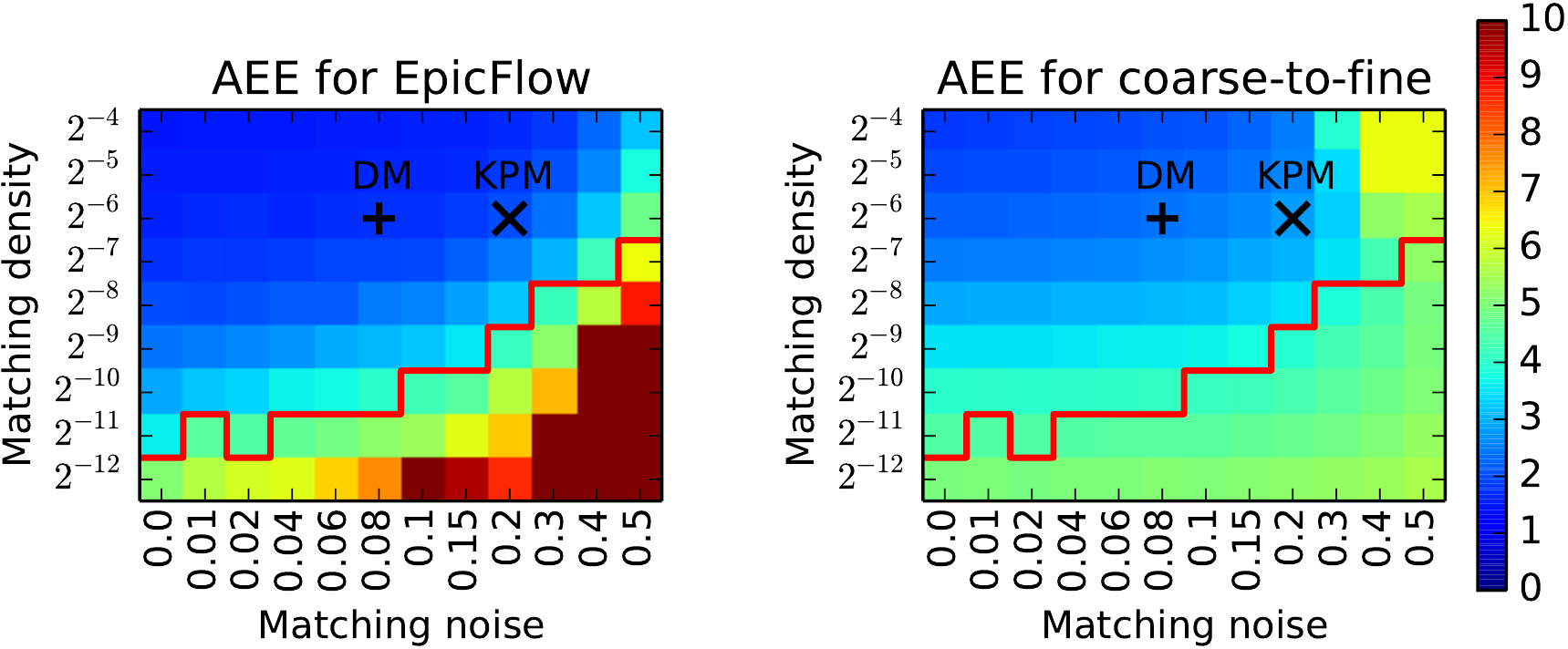}
\caption{Comparison of AEE between EpicFlow~(left) and a coarse-to-fine scheme~(right) for various synthetic input matches with different densities and error levels. 
For positions above the red line, EpicFlow performs better.}
\label{fig:oprange} 

\end{figure}

\noindent \textbf{Sensitivity to the matching quality.}
In order to get a better understanding of why EpicFlow performs better than a coarse-to-fine scheme,
we have evaluated and compared their performances for different densities and error rates of the input matches.
To that aim, we generated synthetic matches by taking the ground-truth flow, removing points in the occluded areas,
subsampling to obtain the desired density and corrupting the
matches to the desired percentage of incorrect matches.
For each set of matches with a given density and quality, we have carefully
determined the parameters of EpicFlow and the coarse-to-fine method 
on the MPI-Sintel training subset, and then evaluated them on the remaining training images. 

Results in term of AEE are given in Figure~\ref{fig:oprange}, where
density is represented vertically as the ratio of \#matches / \#non-occluded pixels
and matching error is represented horizontally as the ratio of \#false matches / \#matches.
We can observe that EpicFlow yields better results provided that the matching is sufficiently dense for a given error rate.   
For low-density or strongly corrupted matches, EpicFlow yields unsatisfactory performance
(Figure~\ref{fig:oprange}~left), while the coarse-to-fine method remains
relatively robust (Figure~\ref{fig:oprange}~right). 
This shows that our interpolation-based heuristic for initializing the flow 
takes better advantage of the input matches than a coarse-to-fine schemes 
for sufficiently dense matches and is able to recover from matching failures.
We have indicated the position of DeepMatching and KPM in terms of density and quality on the plots: 
they lie inside the area in which EpicFlow outperforms a coarse-to-fine scheme.

\subsection{Comparison with the state of the art}
\label{sec:xptest}

Results on MPI-Sintel test set are given in
Table~\ref{tab:sintel}. Parameters are optimized on the MPI-Sintel
training set.  EpicFlow outperforms the state of the art with a gap of
$0.5$ pixel in AEE compared to the second best performing method,
TF+OFM ~\cite{kennedyoptical}, and $1$ pixel compared to the third one,
DeepFlow~\cite{DeepFlow}.  In particular, we improve for both AEE on
occluded areas and AEE over all pixels and for all displacement
ranges. In addition, our approach is significantly faster than most of
the methods, \eg an order of magnitude faster than the second
best.   

\begin{table}
\resizebox{\linewidth}{!}{
\centering %
\begin{tabular}{|l||c|c|ccc||r|}
\hline
 Method & AEE  & AEE-occ & s0-10  & s10-40  & s40+ & Time \\
\hline 
\textbf{EpicFlow}  & \textbf{6.285}  & \textbf{32.564} & \textbf{1.135}  & \textbf{3.727} & \textbf{38.021} & \textbf{16.4s} \\
TF+OFM~\cite{kennedyoptical}    & 6.727 & 33.929 & 1.512 & 3.765 & 39.761 & $\sim$400s \\
DeepFlow~\cite{DeepFlow}  & 7.212  & 38.781 & 1.284  & 4.107  & 44.118  & 19s\\
S2D-Matching~\cite{Leordeanu2013}  & 7.872 & 40.093 & 1.172  & 4.695  & 48.782 & $\sim$2000s \\
Classic+NLP~\cite{sun2014}  & 8.291  & 40.925 & 1.208  & 5.090  & 51.162  & $\sim$800s \\
MDP-Flow2~\cite{mdpof}  & 8.445 & 43.430 & 1.420  & 5.449  & 50.507 & 709s\\
NLTGV-SC~\cite{ranftl2014non} & 8.746 & 42.242 & 1.587 & 4.780 & 53.860 & \\
LDOF~\cite{Bro11a}  & 9.116  & 42.344 & 1.485  & 4.839  & 57.296 & 30s\\
\hline
\end{tabular}}
\caption{Results on MPI-Sintel test set (final version). AEE-occ is the AEE on occluded areas. s0-10 is the AEE
for pixels whose motions is between 0 and 10 px and similarly
for s10-40 and s40+.}
\label{tab:sintel} 
\end{table}

Table~\ref{tab:kitti} reports the results on the Kitti test set for
methods that do not use epipolar geometry or stereo vision. Parameters are optimized on the Kitti training set. 
We can see that EpicFlow performs best in terms of AEE on non-occluded areas. In term
of percentage of erroneous pixels, our method is competitive with the
other algorithms.
When comparing the methods on both Kitti and MPI-Sintel, 
we outperform TF+OFM~\cite{kennedyoptical} and DeepFlow~\cite{DeepFlow} (second and third on MPI-Sintel) on the Kitti dataset, in particular for occluded areas.
We perform on par with NLTGV-SC~\cite{ranftl2014non} on Kitti that we outperform by 2.5 pixels on MPI-Sintel.

\begin{table}
\resizebox{\linewidth}{!}{\centering %
\begin{tabular}{|l||c|c||c|c||r@{ }l|}
\hline
Method  & AEE-noc  & AEE  & Out-Noc 3  & Out-All 3  & Time & \\
\hline
\textbf{EpicFlow}  & \textbf{1.5}  & 3.8  & 7.88\%  & 17.08\% & 16s &\\
NLTGV-SC~\cite{ranftl2014non}  & 1.6 & 3.8 & \textbf{5.93\%} & 11.96\% & 16s & (GPU)\\
BTF-ILLUM~\cite{demetz2014learning} & \textbf{1.5} & \textbf{2.8} & 6.52\% & \textbf{11.03\%} & 80s & \\
TGV2ADCSIFT~\cite{Braux-Zin_2013_ICCV}  & \textbf{1.5}  & 4.5  & 6.20\%  & 15.15\% & 12s & (GPU)\\
Data-Flow~\cite{Vogel2013}  & 1.9  & 5.5  & 7.11\%  & 14.57\% & 180s & \\
DeepFlow~\cite{DeepFlow}  & \textbf{1.5}  & 5.8  & 7.22\%  & 17.79\% & 17s & \\
TF+OFM~\cite{kennedyoptical} & 2.0 & 5.0 & 10.22\% & 18.46\% & 350s & \\
\hline
\end{tabular}}
\caption{Results on Kitti test set. AEE-noc is the AEE over
non-occluded areas. Out-Noc 3 (resp.\ Out-all 3) refers to the percentage
of pixels where flow estimation has an error above 3 pixels in non-occluded
areas (resp.\ all pixels). \newline
}
\label{tab:kitti} 
\end{table}

On the Middlebury test set,
we obtain an AEE below $0.4$ pixel. This is
competitive with the state of the art.  
In this dataset, there are no large displacements, and consequently,
the benefits of a matching-based approach are limited. Note that we have
slightly increased the number of fixed point iterations to 25 in the
variational method for this dataset (still using one level)  in order
to get an additional smoothing effect. This leads to a
gain of $0.1$ pixels (measured on the Middlebury training set when
setting the parameters on MPI-Sintel training set).

\noindent \textbf{Timings.}
While most methods often require several minutes to run on a single
image pair, ours runs in 16.4 seconds for a MPI-Sintel image pair
($1024\times436$ pixels) on one CPU-core at 3.6Ghz. 
In detail, computing DeepMatching takes 15s, extracting SED edges 0.15s, dense interpolation 0.25s, and  
variational minimization 1s. We can observe that 91\% of the time
is spent on matching. 

\begin{figure}
 \centering
 \begin{tabular}{cccc}
 \begin{turn}{90}{\footnotesize{}{}{}~~~Images}\end{turn}       & \includegraphics[width=0.43\linewidth]{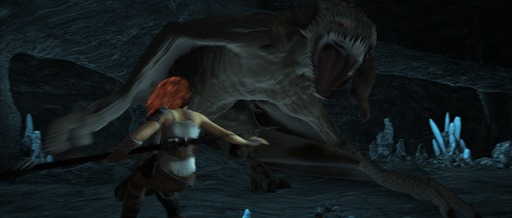}     & \includegraphics[width=0.43\linewidth]{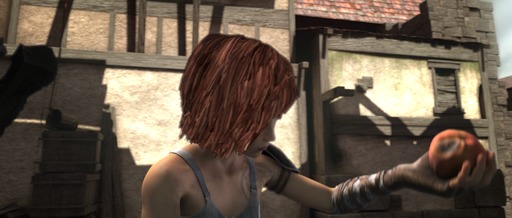} \\
 \begin{turn}{90}{\footnotesize{}{}{}~~~~~~~~GT}\end{turn} & \includegraphics[width=0.43\linewidth]{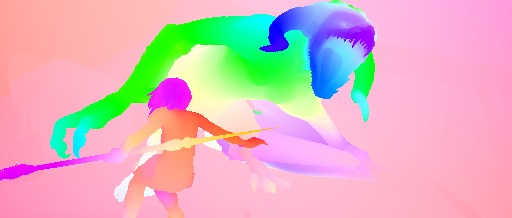}       & \includegraphics[width=0.43\linewidth]{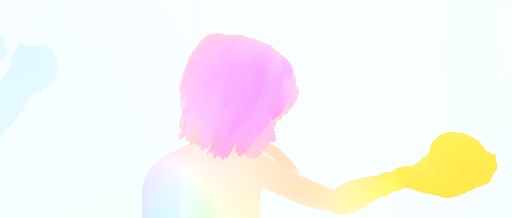} \\
 \begin{turn}{90}{\footnotesize{}{}{}~~EpicFlow}\end{turn}     & \includegraphics[width=0.43\linewidth]{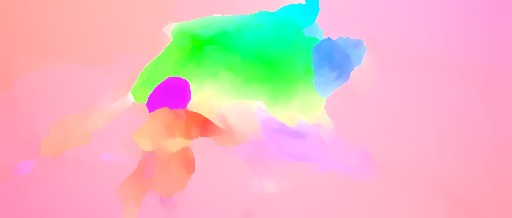} & \includegraphics[width=0.43\linewidth]{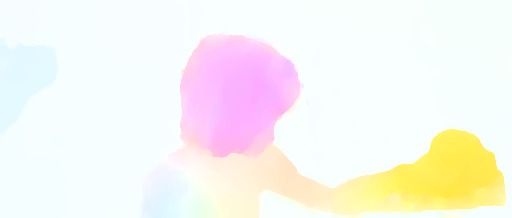}
 \end{tabular}
 \caption{Failure cases of EpicFlow due to missing matches on 
   spear and horns of the dragon (left column) and missing contours
   on the arm (right column).}
 \label{fig:fail}
\end{figure}

\noindent \textbf{Failure cases.}
EpicFlow can be incorrect due to errors in the sparse matches or errors in the contour extraction.  
Figure~\ref{fig:fail} (left column) shows an example where matches
are missing on thin elements (spear and horns of the dragon). Thus,
the optical flow takes the
value of the surrounding region for these elements. 
An example for incorrect contour extraction is presented in
Figure~\ref{fig:fail} (right column). The contour of the character's
left arm is poorly detected. As a result, the motion of the arm
spreads into the background.

\section{Conclusion}

This paper introduces EpicFlow, a novel state-of-the-art optical flow estimation method.
EpicFlow computes a dense correspondence field by performing a sparse-to-dense interpolation 
from an initial sparse set of matches, leveraging contour cues using an edge-aware geodesic distance. 
The approach builds upon the assumption that contours often coincide with motion discontinuities.  
The resulting dense correspondence field is fed as an initial optical flow estimate to a one-level variational energy minimization.  
Experimental results show that EpicFlow outperforms current coarse-to-fine approaches. 
Both the sparse set of matches and the contour estimates are key to our approach.  
Future work will focus on improving these two components separately as well as in an interleaved
manner. 

\section*{Acknowledgments.} This work was supported
by the European integrated project AXES, the MSR-Inria joint centre, 
the LabEx Persyval-Lab (ANR-11-LABX-0025), the Moore-Sloan Data Science Environment at NYU,
and the ERC advanced grant ALLEGRO.

{\small
\bibliographystyle{ieee}
\bibliography{biblio}
}

\end{document}